\documentclass[letterpaper, 10 pt, conference]{ieeeconf}  

\IEEEoverridecommandlockouts                              

\usepackage[hidelinks]{hyperref}
\usepackage[noadjust]{cite}
\usepackage{bm}
\usepackage{bbm}
\usepackage{amsmath}
\usepackage{amssymb}
\usepackage{gensymb}
\usepackage{graphicx}
\DeclareMathOperator*{\argmin}{\arg\!\min}

\DeclareMathOperator*{\expect}{{\huge \mathbb{E}}}
\usepackage{subcaption}
\usepackage{xcolor}
\usepackage{multibib}

\newcommand\Tstrut{\rule{0pt}{2.6ex}}         
\newcommand\Bstrut{\rule[-0.9ex]{0pt}{0pt}}   

\newcites{supp}{Supplementary References}

\title{\LARGE \bf
Learning Reward Functions for Robotic\\Manipulation by Observing Humans
}

\author{Minttu Alakuijala$^{1,2}$, Gabriel Dulac-Arnold$^{3}$, Julien Mairal$^{2}$, Jean Ponce$^{1,4}$ and Cordelia Schmid$^{3}$
\thanks{ \hspace{-0.2cm}
$^{1}$Département d'informatique de l'Ecole normale sup\'{e}rieure (ENS-PSL, \newline \indent CNRS, Inria)
\tt\small \fontsize{8.5}{8.5}\selectfont
\{minttu.alakuijala,jean.ponce\}@inria.fr
}
\thanks{ \hspace{-0.2cm}
$^{2}$Univ. Grenoble Alpes, Inria, CNRS, Grenoble INP, LJK, 38000 \newline \indent Grenoble, France
\tt\small  \fontsize{8.5}{8.5}\selectfont julien.mairal@inria.fr
}%
\thanks{ \hspace{-0.2cm}
$^{3}$Google Research
{\tt\small \fontsize{8.5}{8.5}\selectfont \{dulacarnold,cordelias\}@google.com}
}%
\thanks{ \hspace{-0.2cm}
$^{4}$Courant Institute of Mathematical Sciences and Center for Data Science, \newline \indent New York University
}
}

\begin{document}

\maketitle
\thispagestyle{empty}
\pagestyle{empty}

\begin{abstract}

Observing a human demonstrator manipulate objects provides a rich, scalable and inexpensive source of data for learning robotic policies. However, transferring skills from human videos to a robotic manipulator poses several challenges, not least a difference in action and observation spaces. In this work, we use unlabeled videos of humans solving a wide range of manipulation tasks to learn a task-agnostic reward function for robotic manipulation policies. Thanks to the diversity of this training data, the learned reward function sufficiently generalizes to image observations from a previously unseen robot embodiment and environment to provide a meaningful prior for directed exploration in reinforcement learning. We propose two methods for scoring states relative to a goal image: through direct temporal regression, and through distances in an embedding space obtained with time-contrastive learning. By conditioning the function on a goal image, we are able to reuse one model across a variety of tasks. Unlike prior work on leveraging human videos to teach robots, our method, Human Offline Learned Distances (HOLD) requires neither a priori data from the robot environment, nor a set of task-specific human demonstrations, nor a predefined notion of correspondence across morphologies, yet it is able to accelerate training of several manipulation tasks on a simulated robot arm compared to using only a sparse reward obtained from task completion.

\end{abstract}

\section{Introduction}

Deep learning has greatly advanced the state of the art in applications ranging from computer vision~\cite{he2016deep, dosovitskiy2020image} to natural language processing~\cite{brown2020language, chowdhery2022palm} to speech recognition~\cite{he2019streaming}, but its significance in robotics has been blunted by limited access to large-scale data. Although previous efforts have covered a specific embodiment and task~\cite{levine2018learning,fang2020graspnet}, collecting a massive dataset for each robot and environment of interest is simply not feasible due to the cost of maintenance, human oversight, hardware wear and tear and the bottleneck of real-time execution. For these reasons, creative reuse of data is of central importance for unlocking the benefits of large-scale data-driven learning in robotics.

One potential source of external data is videos of humans performing arbitrary tasks, widely available on the internet and inexpensive to produce. We focus on manipulation tasks in this work, with the aim of learning from crowd-sourced videos of human arms and hands. However, replicating the demonstrated actions and object interactions with a robot is a challenging open problem. On the perception side, there is a significant visual domain gap between observations of a person and of a robot. Human and robot arms usually have very different morphologies and dynamics, particularly in the end-effector, creating a physical domain gap and making a 1:1 mapping between poses ill-defined in general. Moreover, the actions taken by 
humans are not observed unless explicitly recorded with specialized equipment, and hence conventional imitation learning \cite{pomerleau1991efficient,ho2016generative} or offline reinforcement learning \cite{kumar2019stabilizing,fujimoto2019off} methods are not applicable.\\

To overcome these challenges, we investigate the use of videos of people solving manipulation tasks to learn a notion of distance between images from the observation space of a task. We leverage this learned distance as a reward signal on tasks with similar structure but very different visual appearance on a set of robotic manipulation domains that the model has never observed. By training on diverse human demonstrations, we employ a strategy analogous to domain randomization \cite{tobin2017domain} used in sim-to-real transfer, 
which applies variations to visual and physical simulation parameters at training time so that a real-world robotic task with unknown physical properties is more likely to fall in the training distribution. 
Similarly, when trained with different demonstrators, backgrounds, viewpoints, lightings, objects and tasks, our distance model learns to generalize to a variety of manipulator appearances. Furthermore, several aspects of the task as solved by a human are preserved in the robot workspace. For example, object displacements must respect the laws of physics regardless of the actor.

The learned distance function captures roughly how long it takes for an expert to transition from one state to another, and is therefore closely related to a dense reward function representing task progress that can be optimized with reinforcement learning (RL). Learning dense rewards is especially useful in hard exploration problems where it is straightforward to define a sparse task-completion reward, but laborious and error-prone to specify a well-shaped dense reward.

Instead of model-free RL, reward functions estimating task progress can also be optimized with model predictive control
\cite{chen2021learning,tian2021model}, in which case both a forward model of the environment dynamics and a state-action value function need to be learned, typically from undirected exploration data in the target environment. However, these methods require extensive a priori data collection with sufficient coverage on a target robot environment and its action space, and learning accurate video prediction models remains a challenging open problem in itself.
We instead propose to learn a state-value function from observation-only data which allows for the reuse of data from different embodiments, and train a policy for the target embodiment with online RL. We empirically show better sample efficiency per task in online training than was required to learn a model in prior work~\cite{chen2021learning}.


Our contributions are as follows:
i) We propose HOLD\footnote{Code and videos are available on \href{https://sites.google.com/view/hold-rewards}{sites.google.com/view/hold-rewards.}}, a global goal-conditioned distance model which removes the need for demonstration task labels and exact alignment between robot tasks and demonstrated tasks required by prior work~\cite{chen2021learning,schmeckpeper2020reinforcement,zakka2022xirl,shao2021concept2robot,bonardi2020learning,yu2018one,jang2022bc}. 
ii) We show generalization of reward functions trained on unconstrained human videos to robot arms of various morphologies and environments, and accelerate training of model-free RL on 5 simulated manipulation tasks by up to 18x by providing shaped rewards in sparse-reward tasks, or even entirely replacing the reward in some tasks.
iii) We show that time-contrastive embeddings~\cite{sermanet2018time} can successfully represent distances for multiple tasks at once despite a high degree of multi-modality in mixed-task training data.
iv) We show HOLD to outperform existing cross-domain imitation \cite{sermanet2018time} and representation learning \cite{nair2022r3m} approaches able to handle mixed-task videos.

\section{Related work}
\label{sec:related_work}

\paragraph{Intermediate representations} Several prior works have addressed learning robotic policies from human videos via intermediate representations such as pose estimation or keypoint tracking~\cite{qin2022dexmv,petrik2021learning,das2021model}. 
In this work, our aim is to advance the capabilities of learning from raw video data, without depending on hand-crafted intermediate representations of human hands or an object database.

\paragraph{Imitation learning}
Our work is also related to imitation learning from observation, although this line of work has mostly addressed the case of demonstrations from the same observation space \cite{ho2016generative,torabi2018behavioral,aytar2018playing,kostrikov2019discriminator}. We instead tackle the more difficult problem of inverse RL from observation under significant observational and dynamical domain shift.

\paragraph{Offline RL}
Similarly to HOLD, Offline RL~\cite{fujimoto2019off,wu2019behavior,wang2020critic, peng2019advantage} also aims to learn a value function from a dataset of existing trajectories. However, our setting is significantly different from the offline RL problem as we do not have access to either the actions or the rewards of the demonstrator in our dataset, nor do we have a forward model of which states are reachable from a given state, 
making temporal difference based methods not applicable.

\paragraph{Mapping methods}
Many methods for learning from videos seek to learn a direct mapping between demonstration videos and robot states and/or actions, such as an inverse model labeling each human transition with an action from the robot action space
\cite{schmeckpeper2020reinforcement}, or an image-to-image translation of a human demonstration to a corresponding robot demonstration \cite{xiong2021learning,li2021meta}. 
By contrast, our method does not assume a precise 1:1 mapping between the observation and action spaces of the human and the robot and can therefore leverage arbitrarily large amounts of human demonstration videos without any manual supervision cost.

\paragraph{Consistency methods}
A line of prior work has proposed to learn domain-invariant features capturing task progress regardless of whether the actor is a human or a robot arm \cite{schmeckpeper2020reinforcement,zakka2022xirl,sermanet2018time} with reward usually defined as distance to a human demonstration \cite{sermanet2018time} or to a goal state \cite{zakka2022xirl} in the feature space. One issue with using geometrical distances is that transition times between states are not symmetrical if the environment includes unidirectional transitions, such as dropping an object or knocking something down. To account for this, we additionally propose a regression-based model which predicts distances as a function of two ordered states. 
Sequence-based objectives such as temporal cycle consistency \cite{zakka2022xirl} are well suited for single-task learning where all trajectories can be aligned along a global task progression, but it is unclear whether these methods would work on data from several tasks.
%
Most existing approaches to learning robotic manipulation from human videos also require either exact overlap between tasks demonstrated by humans and the robot tasks \cite{chen2021learning,schmeckpeper2020reinforcement,zakka2022xirl,shao2021concept2robot} or robot demonstrations for many of the same tasks \cite{chen2021learning,bonardi2020learning,yu2018one,jang2022bc}. As our model is not specialized for any single task and learns from human data only, no robot demonstrations are needed and the target robot task does not need to be strictly included in the training data as long as a goal image is available to specify the new task.

\paragraph{Time-contrastive embeddings} Sermanet \emph{et al.} \cite{sermanet2018time} propose to use distances in an embedding space learned with a time-contrastive objective, but only consider reward learning for a single task, whereas we learn a single multi-task reward model. Moreover, while \cite{sermanet2018time} propose to directly imitate a human demonstration at 1:1 speed, we instead define the task with a goal image from the robot's observation space. As we show experimentally, \cite{sermanet2018time} needs a nearly identical alignment in the initial states, execution speed and cropping between the video and the robot observations, which is a significant limitation. By contrast, our inverse RL approach requires less supervision and allows the robot to potentially outperform the demonstrator, either by executing the task faster or by finding a more optimal trajectory. Concurrently to our work, Ma \emph{et al.} \cite{ma2022vip} use implicit time-contrastive learning to train a task-agnostic visual reward function related to our time-contrastive model. However, their approach also does not consider the potential asymmetry of dynamics that our regression model can represent.

\paragraph{Functional distance} Our work is also related to estimating functional (also called dynamical) distance between states from online \cite{hartikainen2020dynamical} or offline robot data \cite{tian2021model}. 
However, both works use only robot data from the same environment, without transfer of the action or observation spaces. Our approach is instead based on estimating the state-value function of the demonstrated behavior drawn from an unknown action space. 


\section{Human Offline Learned Distances}
\label{sec:dist_learning}
\subsection{Functional distances from observation-only data}
\label{ssec:dist_learning}
We propose to learn about distances in state space by observing humans and using this prior knowledge of environment dynamics to accelerate training of robotic manipulation policies.
Specifically, our goal is to estimate \emph{functional distance} $d(s, g)$ ~\cite{hartikainen2020dynamical,tian2021model}, between an image $s$ of the current state and a goal image $g$, where $s, g \in \mathcal{S}_r$, the set of camera images from the robot's observation space. This metric should correlate with $\delta(s, g)$, the number of time steps it takes for an expert policy $\pi^*$ to reach the goal $g$ from the state $s$:
\begin{align}
\begin{split}
\delta(s, g) = \expect[T | &s_T = g, s_0 = s, a_t \sim \pi^*(s_t,g), \\
&s_{t+1} \sim p(s_t, a_t)], \label{eq:func_dist}
\end{split}
\end{align}
where
$p$ are the transition dynamics of the environment, modeled as a Markov Decision Process (MDP). We assume each image observation fully captures the environment state in order to unambiguously define tasks using goal images. The negated time difference $-\delta(s, g)$ is equal to the value function $V^*$ for an optimal policy $\pi^*$
and a reward of $-1$ per time step until the episode terminates (upon successfully reaching the goal or exceeding a time limit).
However, this is not the only reward that can be optimized to recover $\pi^*$ (Ng \emph{et al.}~\cite{ng1999policy} discuss conditions for policy-invariant reward shaping in the general case). 
Given the original reward $r=-1$
with $V^*=-\delta$, $\pi^*$ is unchanged for the reward $r'(s_t, a_t, s_{t+1}, g) = -| d(s_{t+1},g) - d(s_t, g)|$ with $V^*=-d$ if we assume that $d(g,g)=0$, $p$ is deterministic and 
pairwise rankings are preserved:
\begin{align}
\forall 
s, s', g \in \mathcal{S}_r \nonumber, \noindent \delta(s, g) > \delta(s', g) &\implies d(s, g) > d(s', g) \\
\text{and } \noindent \delta(s, g) = \delta(s', g) & \implies d(s, g) = d(s', g). \nonumber
\end{align}
\noindent Although defined in terms of an expert policy $\pi^*$, $\delta(s, g)$, and consequently the functions $d(s, g)$ that preserve its rankings, can be estimated from observation-only data, without access to actions $a$, the expert $\pi^*$, or even its action space, by obtaining self-supervised time deltas without manual annotation. While Tian \emph{et al.} \cite{tian2021model} learn the Q-function corresponding to a related, sparse goal-reaching reward
from offline trajectories from the robot, 
our choice of a state-value function, agnostic to a specific action space, 
allows reuse of
data gathered with different but related morphologies, such as other robots or humans. Strictly speaking, the ability to share the function $\delta$ between human and robot MDPs relies on them being isomorphic \cite{schmeckpeper2020reinforcement}, requiring a 1:1 mapping between the action and observation spaces that preserves dynamics $p$. While this may not fully hold in practice, and the distribution of $\delta$ in human data may not necessarily match the robot's dynamics in absolute terms due to embodiment differences, 
the rankings produced by $d$ can be transferred under fewer assumptions. For example, one embodiment may be twice as fast as the other while still preserving all pairwise rankings.

We assume access to a dataset of $N$ video demonstrations of humans executing a variety of manipulation tasks using approximately shortest paths. In practice, the precise length of time may vary significantly across trials and human demonstrators, and depend on the optimality of the demonstration. Although the absolute length of such time intervals may not be consistent across demonstrators, their relative durations provide a useful learning signal; in order to push an object to the right, one must first approach its current position from the left before starting the pushing maneuver, and not the other way around. We present two methods for learning $d$ on this data.

\paragraph{Direct regression (HOLD-R)}
We assume the demonstrations are optimal and pose the functional distance learning problem as a supervised regression task:
\begin{equation}
\theta^*=\argmin \sum_{i=1}^N \sum_{t=1}^{T_i} \sum_{\delta=1}^{T_i-t} || d_\theta(s^i_t, s^i_{t+\delta}) - \delta ||^2_2
\end{equation}
where $s^i_t$ is the $t$th frame of the $i$th video, $T_i$ is the length of the $i$th video, and $d_\theta$ is a function parameterized by $\theta$ trained to predict $\delta$ from Eq. (\ref{eq:func_dist}). The third summation corresponds to data augmentation allowing any future time step in the video to be considered the goal rather than only the last. 

\paragraph{Time-contrastive embeddings (HOLD-C)} Since directly predicting time intervals is difficult and sensitive to noise, we may also consider learning an embedding space where distances can be defined. We propose to use a single-view time-contrastive objective as in TCN \cite{sermanet2018time}. Frames within a small temporal window are encouraged to lie close together in embedding space, whereas embeddings for frames outside some temporal neighborhood are pushed apart. Specifically, if $s_p$ is a positive instance for anchor $s$, and $s_n$ is a negative instance, for all triplets, we want:
\begin{equation}
||f(s) - f(s_p)||^2_2 + m < ||f(s) - f(s_n)||^2_2
\end{equation}
where the margin $m$ is a hyperparameter. However, unlike the single-task setup proposed in \cite{sermanet2018time}, we train $f$ on multi-task data and show it to accelerate robot learning across tasks. Moreover, our method improves upon TCN in several ways at the policy training stage: i)~HOLD enables the robot to outperform the demonstrations by learning relevant shortcuts through interaction, or by simply moving faster, whereas TCN aims to imitate the human. TCN defines the task using a human video, and minimizes distance to each of its states at 1:1 speed -- although the distances are minimized with RL, the best possible reward is defined as matching the human performance. ii)~HOLD requires less supervision: TCN needs one human trajectory of the full task whereas we use distance to a goal image only and require no task demonstrations. iii)~We use a simpler Euclidean distance 
to define the metric $d(s, g)$ in the space $f$, whereas \cite{sermanet2018time} apply a weighted mixture of squared Euclidean and a Huber-style loss $
d(s_t, g_t) = \alpha ||f(s_t) - f(g_t)||^2_2 + \beta \sqrt{\gamma + ||f(s_t) - f(g_t)||^2_2}
$, requiring two additional hyperparameters to be tuned in an already computationally expensive RL training setup. 

\begin{figure*}[t]
    \centering
    \begin{subfigure}[t]{0.495\textwidth}
    \includegraphics[height=1.25cm]{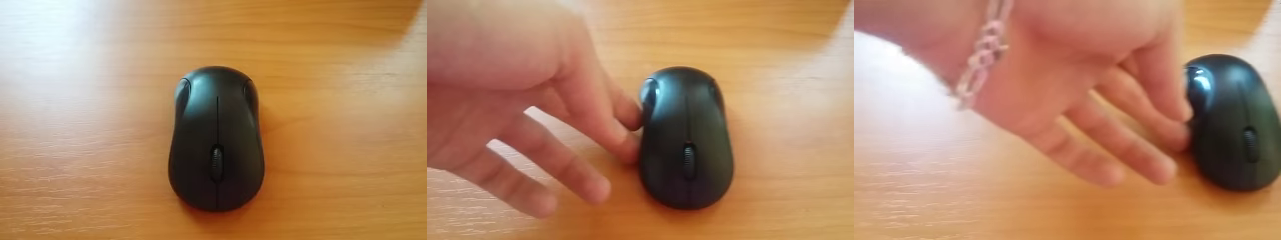}
    \centering
    \caption{Pushing mouse from left to right}
    \end{subfigure}
    \centering
    \begin{subfigure}[t]{0.199\textwidth}
    \includegraphics[height=1.25cm]{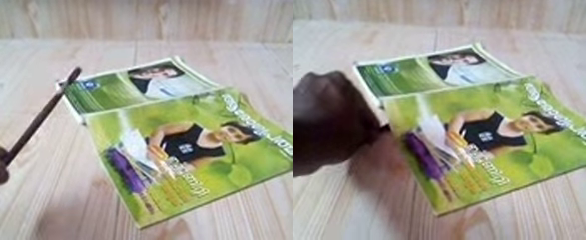}
    \caption{Putting paint brush\\underneath magazine}
    \end{subfigure}
    \centering
    \begin{subfigure}[t]{0.268\textwidth}
    \includegraphics[height=1.25cm]{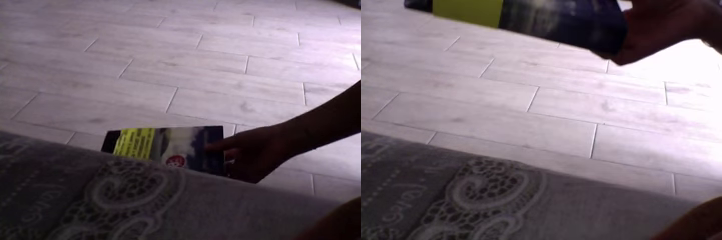}
    \centering
    \caption{Moving book up}
    \end{subfigure}
    \centering
    \caption{Example human videos from Something-Something v2 used to train the distance models.}
    \label{fig:ssv2_videos}
\vspace{0.3cm}
    \centering
    \begin{subfigure}[b]{0.138\textwidth}
    \includegraphics[width=\linewidth]{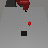}
    \caption{Pushing: start}
    \label{fig:rlv_pushing_initial_state}
    \end{subfigure}
    \centering
    \begin{subfigure}[b]{0.138\textwidth}
    \includegraphics[width=\linewidth]{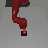}
    \caption{Pushing: goal}
    \label{fig:rlv_pushing_goal}
    \end{subfigure}    \hspace{0.6cm}
    \begin{subfigure}[b]{0.138\textwidth}
    \includegraphics[width=\linewidth]{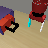}
    \caption{Drawer: start}
    \label{fig:rlv_open_drawer_init}
    \end{subfigure}
    \begin{subfigure}[b]{0.138\textwidth}
    \includegraphics[width=\linewidth]{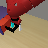}
    \caption{Drawer subtask}
    \label{fig:rlv_open_drawer1}
    \end{subfigure}
    \begin{subfigure}[b]{0.138\textwidth}
    \includegraphics[width=\linewidth]{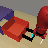}
    \caption{Drawer: goal}
    \label{fig:rlv_open_drawer2}
    \end{subfigure}
    \vspace{0.1cm} \\
    \centering
    \begin{subfigure}[b]{0.179\textwidth}
    \includegraphics[width=\linewidth]{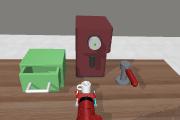}
    \caption{Start state for all tasks}
    \label{fig:dvd_initial_state}
    \end{subfigure}
    \centering
    \begin{subfigure}[b]{0.19\textwidth}
    \includegraphics[width=\linewidth]{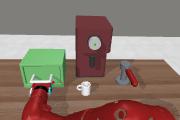}
    \caption{Close Drawer}
    \label{fig:dvd_close_drawer_goal}
    \end{subfigure}
    \begin{subfigure}[b]{0.19\textwidth}
    \includegraphics[width=\linewidth]{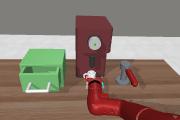}
    \caption{Push Cup Forward}
    \label{fig:push_cup_goal}
    \end{subfigure}
    \begin{subfigure}[b]{0.19\textwidth}
    \includegraphics[width=\linewidth]{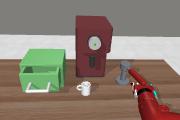}
    \caption{Turn Faucet Right}
    \label{fig:dvd_faucet_right_goal}
    \end{subfigure}
    \caption{(a-e) The RLV Tasks. (f-i) The DVD tasks: a Sawyer arm in a tabletop environment adapted from Meta-World~\cite{yu2020meta}.}
    \label{fig:all_tasks}
\end{figure*}

\subsection{Policy learning}
\label{ssec:policy_learning}
We propose to use the learned functional distance to define a dense reward function for an RL policy. Although our reward function is goal-conditioned and shared across tasks, we train one policy per robot task, in order to focus on multi-task reward learning in this work.
As we want to minimize distance to the goal frame, we define reward as follows:
\begin{equation}
r(s_t, a_t, s_{t+1}, g) = -\max(0, d(s_{t+1},g)-d(g, g)) / T
\label{eq:func_dist_reward}
\end{equation}
where $s_t, s_{t+1}, g \in \mathcal{S}_r$, $a_t$ is an action from the robot's action space, and $T$ is an optional normalizer. We subtract the baseline $d(g, g)$ from the distance estimates to ensure arriving at the goal has reward 0 and no other state has higher reward; $d(g, g)$ may be positive for the regression models due to untrimmed training videos where the demonstrator idles after solving the task. This definition of reward corresponds to minimizing the sum of distances until the end of the episode, as done by \cite{tian2021model,hartikainen2020dynamical}. Alternatively, $r$ could be defined based on the difference $d(s_{t+1}, g) - d(s_t, g)$, such that only the reduction is maximized for each time step. However, we found the cumulative form to perform better empirically, possibly due to being less noisy.


	

\section{Experimental Results}
\label{sec:results}
\subsection{Distance learning}
\paragraph{Dataset}
We train HOLD on Something-Something v2 (SSv2) \cite{goyal2017something}, a crowd-sourced dataset of 220,847 video clips of 174 action classes (with examples in Fig.~\ref{fig:ssv2_videos}). Each action is demonstrated with arbitrary objects, matching templates such as \emph{Moving [something] closer to [something]}. The clips last 4 seconds on average and are mostly filmed using handheld devices, with non-negligible camera motion. Although SSv2 videos are grouped into discrete action classes, we do not make use of these labels\footnote{Only HOLD-R with ViViT architecture made use of labels in pretraining, whereas HOLD-R with ResNet backbones as well as all HOLD-C models did not. However, the pretraining could have potentially been done on a different labeled dataset such as Kinetics~\cite{kay2017kinetics} or skipped altogether, and no labels are needed for the regression task.}, 
making our method applicable on any large-scale goal-oriented manipulation data. As we train a single goal image-conditioned distance function, there also does not need to be exact overlap between the demonstrated tasks and the target tasks on the robot, unlike in prior works \cite{chen2021learning,schmeckpeper2020reinforcement,zakka2022xirl,shao2021concept2robot,bonardi2020learning,yu2018one,jang2022bc}.

\paragraph{Training details}
We consider two sizes of network architecture: a ResNet-50~\cite{he2016deep} and a Video Vision Transformer (ViViT)~\cite{arnab2021vivit} pretrained on SSv2 classification. As the single-view time-contrastive objective only supports embedding single images, for HOLD-C we instead use either a ResNet or a Vision Transformer (ViT)~\cite{dosovitskiy2020image} pretrained on ImageNet-21K. 
We train the ResNet models from scratch, and fine-tune the pretrained models on SSv2 without labels after replacing their classification heads. To adapt the pretrained ViViT model for regression, we also reinitialize its temporal position embeddings and shorten the temporal window to 4, including the 3 most recent frames and one goal frame. We also reduce the temporal filter dimension to 1 as there is no longer a computational benefit to shortening the sequence length. For time-contrastive training, we sample batches of 32 subsequent frames per video and use a positive window of 0.2 seconds and a negative window of 0.4s, as done by \cite{sermanet2018time}. For both objectives, we apply the same data augmentation procedure as \cite{arnab2021vivit}, but leave out MixUp. For further training details, see Appendix \ref{app:training_details}. 
We observed better policy training performance for the ResNet model for HOLD-C, and for ViViT for HOLD-R, so we report results using these backbones in Section \ref{ssec:policy_learning_results}. Ablations using the other architectures are included in Appendix \ref{app:distance_model_ablations}, and strategies for evaluating the distance models on human data before testing them in robot policy training are discussed in Appendix \ref{app:spearman}.


\subsection{Policy Learning}
\label{ssec:policy_learning_results}
To demonstrate the utility of our method as a reward function for training RL policies, we evaluate it on the Pushing and Drawer Opening tasks from RLV \cite{schmeckpeper2020reinforcement} (Fig. \ref{fig:rlv_pushing_initial_state}--\ref{fig:rlv_open_drawer2}) and on the Close Drawer, Push Cup Forward and Turn Faucet Right tasks from DVD \cite{chen2021learning} (Fig. \ref{fig:dvd_initial_state}--\ref{fig:dvd_faucet_right_goal}). We follow prior work \cite{schmeckpeper2020reinforcement,zakka2022xirl} in using Soft Actor-Critic (SAC)~\cite{haarnoja2018soft} as the underlying RL algorithm and evaluate it on 20 episodes for all tasks. All policies use images as input, and we reuse the policy and critic architectures as well as algorithm hyperparameters from \cite{schmeckpeper2020reinforcement}.
Like \cite{schmeckpeper2020reinforcement}, we augment our learned reward from Eq. (\ref{eq:func_dist_reward}) with a sparse task reward: 1 for success, and 0 otherwise, defined by each environment based on distance to the target configuration. Since the predicted distances can be significantly larger than 1 but should not override the sparse reward, we scale the predicted rewards by $1/T$, where $T$ is set such that the scale of initial state distances is $\approx 1/3$. Ablations for other values are included in Appendix \ref{app:distance_model_ablations}.


\subsubsection{RLV tasks}\label{ssec:rlv_tasks}
\begin{figure*}[t]
    \centering
    \begin{subfigure}[t]{\textwidth}
    \centering
    \includegraphics[height=0.8cm]{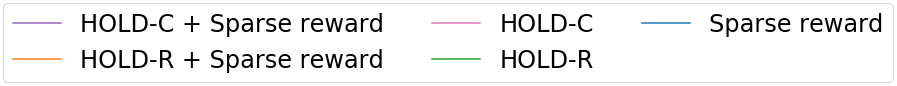}
    \end{subfigure}\\
    \centering
    \begin{subfigure}[b]{0.31\textwidth}
    \includegraphics[width=\linewidth]{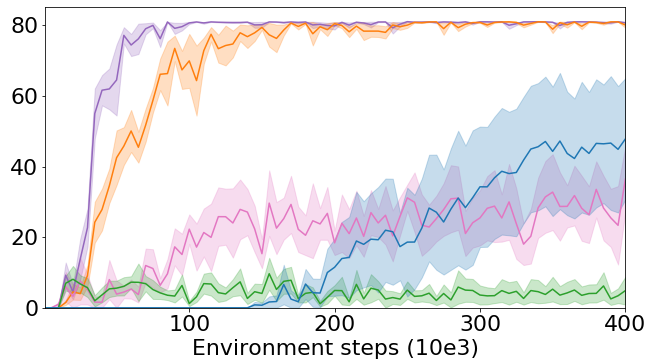}
    \caption{Pushing}
    \label{fig:rlv_pushing_results}
    \end{subfigure}
    \centering
    \begin{subfigure}[b]{0.31\textwidth}
    \includegraphics[width=\linewidth]{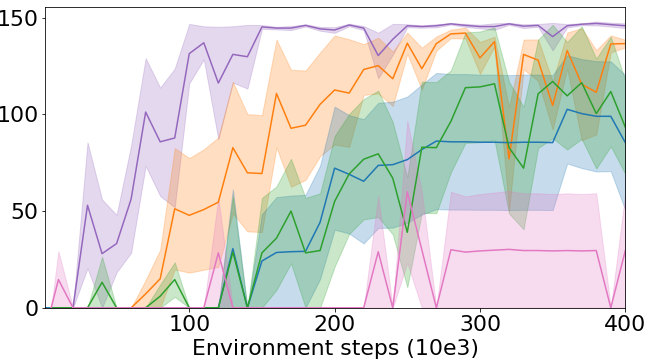}
    \caption{Drawer Opening (2 subtasks)}
    \label{fig:rlv_drawer_results}
    \end{subfigure}
    \caption{Return on the RLV tasks (5 random seeds, with standard error).}
    \label{fig:rlv_results}
    \vspace{0.3cm}
    \centering
    \begin{subfigure}[t]{\textwidth}
    \centering
    \includegraphics[height=0.8cm]{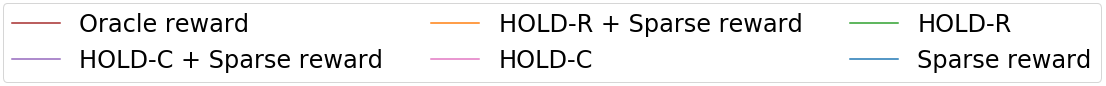}
    \end{subfigure}
    \centering
    \begin{subfigure}[t]{0.31\textwidth}
    \includegraphics[width=\linewidth]{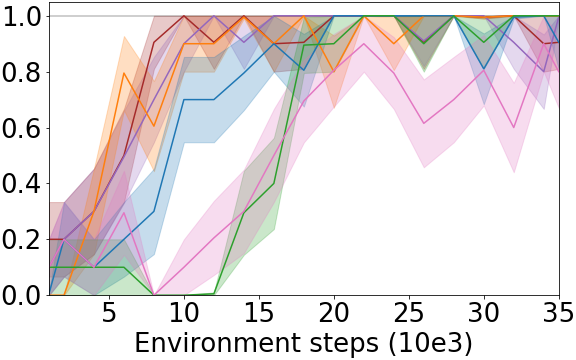}
    \subcaption{Close Drawer}
    \label{fig:dvd_drawer_results}
    \end{subfigure}
    \centering
    \begin{subfigure}[t]{0.31\textwidth}
    \includegraphics[width=\linewidth]{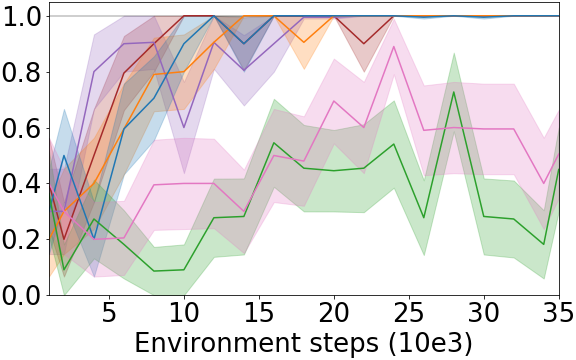}
    \subcaption{Push Cup Forward}
    \label{fig:dvd_cup_results}
    \end{subfigure}
    \begin{subfigure}[t]{0.31\textwidth}
    \includegraphics[width=\linewidth]{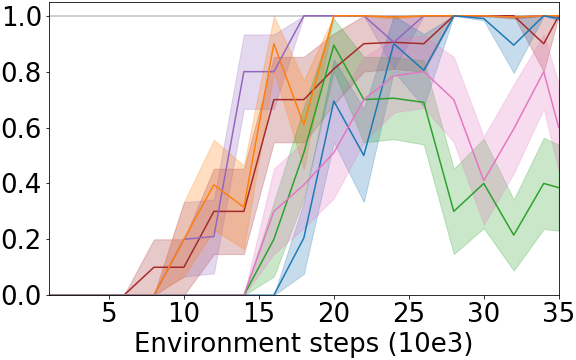}
    \subcaption{Turn Faucet Right}
    \label{fig:dvd_faucet_results}
    \end{subfigure}
    \centering
    \caption{Success rates on the DVD tasks (10 random seeds, with standard error). Our reward functions improve over sparse reward, and learn the Close Drawer task without using sparse reward.}
    \label{fig:dvd_results}
\end{figure*}
As shown in Fig. \ref{fig:rlv_results}, the sum of both reward functions, appropriately balanced, significantly accelerates training compared to using the sparse reward alone. In our experiments, using only sparse reward required 10x more samples for Pushing and $>$18x more for Drawer to reach the return of \mbox{HOLD-C}.
We find that HOLD-C outperforms HOLD-R for Pushing, both with and without sparse reward, and in the sparse reward setting for Drawer Opening.

\paragraph{Pushing} Without added sparse reward, a single failure case is prominent: while the policy quickly learns to match the end-effector position in the goal frame, it fails to pay attention to the puck position. As observed by Tian \emph{et al.} \cite{tian2021model}, it is easy for the distance function to excessively focus on fully actuated components in the scene as these are highly predictive of temporal offset. Although HOLD is able to generalize from human arms to a robot arm, 
for tasks with variable object positions, it may 
be better suited as an exploration strategy used together with an otherwise rarely-observed sparse reward than a standalone multi-task reward. Note that although Zakka \emph{et al.} \cite{zakka2022xirl} also evaluate on Pushing, their results are not comparable as their method is trained on the easier RLV Pushing dataset \cite{schmeckpeper2020reinforcement} collected to match the appearance of the robot task, and report on the simpler State Pusher task where the policy directly observes the 2D puck position and the 3D end-effector position.

\paragraph{Drawer} The Drawer Opening task has double the episode length (200 steps) of Pushing, and consists of two distinct motions: approaching and inserting the gripper into the handle, and pulling the drawer open once there. We find that applying the HOLD models on the full task suffers from the local minimum of only matching the arm position in the goal image. However, if we instead define the task in two parts using an intermediate goal image (in Fig. \ref{fig:rlv_open_drawer1}), our rewards significantly improve sample efficiency compared to the sparse task-completion reward provided by the environment alone, as shown in Fig. \ref{fig:rlv_drawer_results}. Moreover, HOLD-R alone without any environment reward performs on par with sparse reward in this setting. We train a single policy for both subtasks, which is conditioned on the active goal image by concatenating it to the observation $s$. For all distance functions, we switch to the next subtask when $d<1$ for at least 3 consecutive time steps.

\subsubsection{DVD tasks}\label{ssec:dvd_tasks}
We report success rate for the DVD tasks in Fig. \ref{fig:dvd_results}. These tasks are significantly easier than the RLV tasks and quickly learned even using only sparse reward. To estimate the upper bound in learning speed achievable by improving the reward alone, we define an oracle reward using knowledge of robot and object positions. Since we observe only a narrow performance gap between the oracle and the sparse reward, the learning speed in these tasks is limited mostly by the RL algorithm. Although it is therefore difficult to show much improvement over the sparse reward, both HOLD models outperform it, particularly for Close Drawer and Turn Faucet Right. For Close Drawer, HOLD also solves the task without sparse reward. Unlike \cite{chen2021learning}, we do not first collect a dataset of 10,000 trajectories, or 600,000 steps, of random exploration on the robot 
to learn a model of the environment, but instead focus on the model-free setting. 
We show adaptation to a new robot, set of objects and environment in just 12,000--18,000 steps, or 200 to 300 episodes, when sparse environment reward is available, or 22,000 without sparse reward for Close Drawer.

\begin{figure*}
\centering
    \begin{subfigure}[t]{\textwidth}
    \centering
    \includegraphics[height=0.8cm]{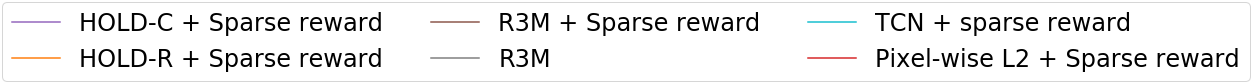}
    \end{subfigure}
    \begin{subfigure}[b]{0.31\textwidth}
    \includegraphics[width=\linewidth]{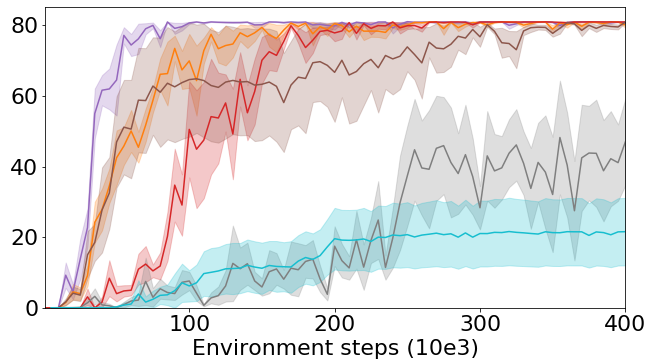}
    \caption{Pushing}
    \label{fig:rlv_pushing_baseline_results}
    \end{subfigure}
    \begin{subfigure}[b]{0.31\textwidth}
    \includegraphics[width=\linewidth]
    {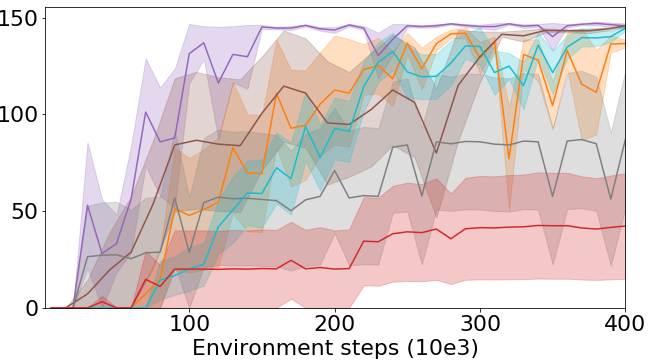}
    \caption{Drawer Opening (2 subtasks)}
    \label{fig:rlv_drawer_baseline_results}
    \end{subfigure}
\caption{HOLD-C outperforms TCN and R3M rewards on both RLV tasks.}
\label{fig:rlv_baseline_results}
%
\end{figure*}

\subsection{Baseline comparisons}\label{ssec:sota_comparison}
We compare HOLD to rewards defined by two prior methods: TCN \cite{sermanet2018time} and R3M \cite{nair2022r3m}. We note that single-task learning methods RLV~\cite{schmeckpeper2020reinforcement} and XIRL~\cite{zakka2022xirl} are not applicable to our setting as they require demonstration task labels. TCN proposes to transfer a policy given a human video demonstration by minimizing distance to the embeddings of each of the visited states $g_t$ in turn. 
We empirically set the hyperparameters of $d(s_t, g_t)$ to $\alpha=0.005, \beta=0.02$, and $\gamma=0.2$. As performance may vary based on the exact demonstration video used, we evaluate 3 demonstrations per task from the RLV dataset, which is collected to closely match the RLV robot tasks, and report the average performance across demonstrations (trained with 5 seeds each) in Figure \ref{fig:rlv_baseline_results}. Even the closely aligned demonstrations transfer poorly to policy learning, especially for Pushing, due to slight differences in initial state, cropping or execution speed, highlighting the brittleness of the trajectory-following objective of TCN.

Although R3M is proposed as a general feature representation, we also compare against using Euclidean distance in the representation space for defining dense rewards. We used the ResNet-50 model checkpoint from \cite{nair2022r3m}, trained on the much larger Ego4D \cite{grauman2022ego4d} (3,500 hours) rather than SSv2 (200~hours). As shown in Figure \ref{fig:rlv_baseline_results}, HOLD-C outperforms R3M in both RLV tasks despite having been trained on less data and requiring no language descriptions. Like our method, R3M also requires sparse rewards to fully solve the tasks, and an intermediate goal for Drawer opening.

We also include a simple baseline of using the negative pixel-wise distance in image space $-||s - g||_2$ as a reward. While this baseline with sparse reward also learns Pushing faster than sparse reward alone, as shown in Figure \ref{fig:rlv_baseline_results}, it still requires many more training samples than either HOLD-R or HOLD-C, and fails to reliably learn Drawer.

\section{Conclusion}

\label{sec:conclusion}
We have presented a method for learning goal image conditioned reward functions for robotic manipulation from unlabeled human videos, in a challenging setting which no prior work has addressed to our knowledge. Learning a prior for robot behavior from a dataset of human demonstrations without task labels requires generalization both across tasks and across a significant domain shift. While 
most accurate for short-horizon tasks,
the distance functions we train produce useful rewards for visually different robot environments that are able to accelerate training over using sparse reward alone, and can be composed to perform more general multi-step manipulation tasks using subgoals. Finally, we have shown that for some tasks, the predicted rewards alone are sufficient to learn the task without any additional success signals.

\section*{Acknowledgements}
This work was in part suppported by the Inria / NYU
collaboration, the Louis Vuitton / ENS chair on artificial
intelligence and the French government under management
of Agence Nationale de la Recherche as part of the \emph{Investissements d'avenir} program (PRAIRIE 3IA Institute) as well as under ANR 3IA MIAI@Grenoble Alpes (ANR-19-P3IA-0003).
It was performed using HPC resources from GENCI–IDRIS
(Grant 2022-AD011013362). Minttu Alakuijala was supported in part by a Google
CIFRE PhD Fellowship. We would like to thank Elliot Chane-Sane for reviewing this manuscript.









\bibliographystyle{IEEEtran}
\bibliography{IEEEabrv,root}  

\begin{thebibliography}{10}
\providecommand{\url}[1]{#1}
\csname url@samestyle\endcsname
\providecommand{\newblock}{\relax}
\providecommand{\bibinfo}[2]{#2}
\providecommand{\BIBentrySTDinterwordspacing}{\spaceskip=0pt\relax}
\providecommand{\BIBentryALTinterwordstretchfactor}{4}
\providecommand{\BIBentryALTinterwordspacing}{\spaceskip=\fontdimen2\font plus
\BIBentryALTinterwordstretchfactor\fontdimen3\font minus
  \fontdimen4\font\relax}
\providecommand{\BIBforeignlanguage}[2]{{%
\expandafter\ifx\csname l@#1\endcsname\relax
\typeout{** WARNING: IEEEtran.bst: No hyphenation pattern has been}%
\typeout{** loaded for the language `#1'. Using the pattern for}%
\typeout{** the default language instead.}%
\else
\language=\csname l@#1\endcsname
\fi
#2}}
\providecommand{\BIBdecl}{\relax}
\BIBdecl

\bibitem{he2016deep}
K.~He, X.~Zhang, S.~Ren, and J.~Sun, ``Deep residual learning for image
  recognition,'' in \emph{Proceedings of the IEEE Conference on Computer Vision
  and Pattern Recognition}, 2016, pp. 770--778.

\bibitem{dosovitskiy2020image}
A.~Dosovitskiy, L.~Beyer, A.~Kolesnikov, D.~Weissenborn, X.~Zhai,
  T.~Unterthiner, M.~Dehghani, M.~Minderer, G.~Heigold, S.~Gelly \emph{et~al.},
  ``An image is worth 16x16 words: Transformers for image recognition at
  scale,'' in \emph{International Conference on Learning Representations},
  2021.

\bibitem{brown2020language}
T.~Brown, B.~Mann, N.~Ryder, M.~Subbiah, J.~D. Kaplan, P.~Dhariwal,
  A.~Neelakantan, P.~Shyam, G.~Sastry, A.~Askell \emph{et~al.}, ``Language
  models are few-shot learners,'' \emph{Advances in Neural Information
  Processing Systems}, vol.~33, pp. 1877--1901, 2020.

\bibitem{chowdhery2022palm}
A.~Chowdhery, S.~Narang, J.~Devlin, M.~Bosma, G.~Mishra, A.~Roberts, P.~Barham,
  H.~W. Chung, C.~Sutton, S.~Gehrmann \emph{et~al.}, ``Palm: Scaling language
  modeling with pathways,'' \emph{arXiv preprint arXiv:2204.02311}, 2022.

\bibitem{he2019streaming}
Y.~He, T.~N. Sainath, R.~Prabhavalkar, I.~McGraw, R.~Alvarez, D.~Zhao,
  D.~Rybach, A.~Kannan, Y.~Wu, R.~Pang \emph{et~al.}, ``Streaming end-to-end
  speech recognition for mobile devices,'' in \emph{IEEE International
  Conference on Acoustics, Speech and Signal Processing (ICASSP)}, 2019, pp.
  6381--6385.

\bibitem{levine2018learning}
S.~Levine, P.~Pastor, A.~Krizhevsky, J.~Ibarz, and D.~Quillen, ``Learning
  hand-eye coordination for robotic grasping with deep learning and large-scale
  data collection,'' \emph{The International journal of robotics research},
  vol.~37, no. 4-5, pp. 421--436, 2018.

\bibitem{fang2020graspnet}
H.-S. Fang, C.~Wang, M.~Gou, and C.~Lu, ``Graspnet-1billion: A large-scale
  benchmark for general object grasping,'' in \emph{Proceedings of the IEEE/CVF
  Conference on Computer Vision and Pattern Recognition}, 2020, pp.
  11\,444--11\,453.

\bibitem{pomerleau1991efficient}
D.~A. Pomerleau, ``Efficient training of artificial neural networks for
  autonomous navigation,'' \emph{Neural computation}, vol.~3, no.~1, pp.
  88--97, 1991.

\bibitem{ho2016generative}
J.~Ho and S.~Ermon, ``Generative adversarial imitation learning,''
  \emph{Advances in Neural Information Processing Systems}, vol.~29, pp.
  4565--4573, 2016.

\bibitem{kumar2019stabilizing}
A.~Kumar, J.~Fu, M.~Soh, G.~Tucker, and S.~Levine, ``Stabilizing off-policy
  q-learning via bootstrapping error reduction,'' \emph{Advances in Neural
  Information Processing Systems}, vol.~32, 2019.

\bibitem{fujimoto2019off}
S.~Fujimoto, D.~Meger, and D.~Precup, ``Off-policy deep reinforcement learning
  without exploration,'' in \emph{International Conference on Machine
  Learning}, 2019, pp. 2052--2062.

\bibitem{tobin2017domain}
J.~Tobin, R.~Fong, A.~Ray, J.~Schneider, W.~Zaremba, and P.~Abbeel, ``Domain
  randomization for transferring deep neural networks from simulation to the
  real world,'' in \emph{IEEE/RSJ International Conference on Intelligent
  Robots and Systems}, 2017, pp. 23--30.

\bibitem{chen2021learning}
A.~S. Chen, S.~Nair, and C.~Finn, ``Learning generalizable robotic reward
  functions from "in-the-wild" human videos,'' in \emph{RSS}, 2021.

\bibitem{tian2021model}
S.~Tian, S.~Nair, F.~Ebert, S.~Dasari, B.~Eysenbach, C.~Finn, and S.~Levine,
  ``Model-based visual planning with self-supervised functional distances,'' in
  \emph{International Conference on Learning Representations}, 2021.

\bibitem{schmeckpeper2020reinforcement}
K.~Schmeckpeper, O.~Rybkin, K.~Daniilidis, S.~Levine, and C.~Finn,
  ``Reinforcement learning with videos: Combining offline observations with
  interaction,'' in \emph{Conference on Robot Learning}, 2020.

\bibitem{zakka2022xirl}
K.~Zakka, A.~Zeng, P.~Florence, J.~Tompson, J.~Bohg, and D.~Dwibedi, ``Xirl:
  Cross-embodiment inverse reinforcement learning,'' in \emph{Conference on
  Robot Learning}.\hskip 1em plus 0.5em minus 0.4em\relax PMLR, 2022, pp.
  537--546.

\bibitem{shao2021concept2robot}
L.~Shao, T.~Migimatsu, Q.~Zhang, K.~Yang, and J.~Bohg, ``Concept2robot:
  Learning manipulation concepts from instructions and human demonstrations,''
  \emph{The International Journal of Robotics Research}, vol.~40, no. 12-14,
  pp. 1419--1434, 2021.

\bibitem{bonardi2020learning}
A.~Bonardi, S.~James, and A.~J. Davison, ``Learning one-shot imitation from
  humans without humans,'' \emph{IEEE Robotics and Automation Letters}, vol.~5,
  no.~2, pp. 3533--3539, 2020.

\bibitem{yu2018one}
T.~Yu, C.~Finn, A.~Xie, S.~Dasari, T.~Zhang, P.~Abbeel, and S.~Levine,
  ``One-shot imitation from observing humans via domain-adaptive
  meta-learning,'' in \emph{RSS}, 2018.

\bibitem{jang2022bc}
E.~Jang, A.~Irpan, M.~Khansari, D.~Kappler, F.~Ebert, C.~Lynch, S.~Levine, and
  C.~Finn, ``Bc-z: Zero-shot task generalization with robotic imitation
  learning,'' in \emph{Conference on Robot Learning}.\hskip 1em plus 0.5em
  minus 0.4em\relax PMLR, 2022, pp. 991--1002.

\bibitem{sermanet2018time}
P.~Sermanet, C.~Lynch, Y.~Chebotar, J.~Hsu, E.~Jang, S.~Schaal, S.~Levine, and
  G.~Brain, ``Time-contrastive networks: Self-supervised learning from video,''
  in \emph{IEEE International Conference on Robotics and Automation}, 2018, pp.
  1134--1141.

\bibitem{nair2022r3m}
S.~Nair, A.~Rajeswaran, V.~Kumar, C.~Finn, and A.~Gupta, ``R3m: A universal
  visual representation for robot manipulation,'' in \emph{Conference on Robot
  Learning}, 2022.

\bibitem{qin2022dexmv}
Y.~Qin, Y.-H. Wu, S.~Liu, H.~Jiang, R.~Yang, Y.~Fu, and X.~Wang, ``Dexmv:
  Imitation learning for dexterous manipulation from human videos,'' in
  \emph{ECCV 2022}.\hskip 1em plus 0.5em minus 0.4em\relax Springer, 2022, pp.
  570--587.

\bibitem{petrik2021learning}
V.~Petr{\'\i}k, M.~Tapaswi, I.~Laptev, and J.~Sivic, ``Learning object
  manipulation skills via approximate state estimation from real videos,'' in
  \emph{Conference on Robot Learning}.\hskip 1em plus 0.5em minus 0.4em\relax
  PMLR, 2021, pp. 296--312.

\bibitem{das2021model}
N.~Das, S.~Bechtle, T.~Davchev, D.~Jayaraman, A.~Rai, and F.~Meier,
  ``Model-based inverse reinforcement learning from visual demonstrations,'' in
  \emph{Conference on Robot Learning}, 2021, pp. 1930--1942.

\bibitem{torabi2018behavioral}
F.~Torabi, G.~Warnell, and P.~Stone, ``Behavioral cloning from observation,''
  in \emph{Proceedings of the 27th International Joint Conference on Artificial
  Intelligence}, 2018, pp. 4950--4957.

\bibitem{aytar2018playing}
Y.~Aytar, T.~Pfaff, D.~Budden, T.~Paine, Z.~Wang, and N.~De~Freitas, ``Playing
  hard exploration games by watching youtube,'' \emph{Advances in Neural
  Information Processing Systems}, vol.~31, 2018.

\bibitem{kostrikov2019discriminator}
I.~Kostrikov, K.~K. Agrawal, D.~Dwibedi, S.~Levine, and J.~Tompson,
  ``Discriminator-actor-critic: Addressing sample inefficiency and reward bias
  in adversarial imitation learning,'' in \emph{International Conference on
  Learning Representations}, 2019.

\bibitem{wu2019behavior}
Y.~Wu, G.~Tucker, and O.~Nachum, ``Behavior regularized offline reinforcement
  learning,'' \emph{arXiv preprint arXiv:1911.11361}, 2019.

\bibitem{wang2020critic}
Z.~Wang, A.~Novikov, K.~Zolna, J.~S. Merel, J.~T. Springenberg, S.~E. Reed,
  B.~Shahriari, N.~Siegel, C.~Gulcehre, N.~Heess \emph{et~al.}, ``Critic
  regularized regression,'' \emph{Advances in Neural Information Processing
  Systems}, vol.~33, 2020.

\bibitem{peng2019advantage}
X.~B. Peng, A.~Kumar, G.~Zhang, and S.~Levine, ``Advantage-weighted regression:
  Simple and scalable off-policy reinforcement learning,'' \emph{arXiv preprint
  arXiv:1910.00177}, 2019.

\bibitem{xiong2021learning}
H.~Xiong, Q.~Li, Y.-C. Chen, H.~Bharadhwaj, S.~Sinha, and A.~Garg, ``Learning
  by watching: Physical imitation of manipulation skills from human videos,''
  in \emph{IEEE/RSJ International Conference on Intelligent Robots and
  Systems}, 2021, pp. 7827--7834.

\bibitem{li2021meta}
J.~Li, T.~Lu, X.~Cao, Y.~Cai, and S.~Wang, ``Meta-imitation learning by
  watching video demonstrations,'' in \emph{International Conference on
  Learning Representations}, 2021.

\bibitem{ma2022vip}
Y.~J. Ma, S.~Sodhani, D.~Jayaraman, O.~Bastani, V.~Kumar, and A.~Zhang, ``Vip:
  Towards universal visual reward and representation via value-implicit
  pre-training,'' \emph{arXiv preprint arXiv:2210.00030}, 2022.

\bibitem{hartikainen2020dynamical}
K.~Hartikainen, X.~Geng, T.~Haarnoja, and S.~Levine, ``Dynamical distance
  learning for semi-supervised and unsupervised skill discovery,'' in
  \emph{International Conference on Learning Representations}, 2020.

\bibitem{ng1999policy}
A.~Y. Ng, D.~Harada, and S.~Russell, ``Policy invariance under reward
  transformations: Theory and application to reward shaping,'' in
  \emph{International Conference on Machine Learning}, vol.~99, 1999, pp.
  278--287.

\bibitem{yu2020meta}
T.~Yu, D.~Quillen, Z.~He, R.~Julian, K.~Hausman, C.~Finn, and S.~Levine,
  ``Meta-world: A benchmark and evaluation for multi-task and meta
  reinforcement learning,'' in \emph{Conference on Robot Learning}.\hskip 1em
  plus 0.5em minus 0.4em\relax PMLR, 2020, pp. 1094--1100.

\bibitem{goyal2017something}
R.~Goyal, S.~Ebrahimi~Kahou, V.~Michalski, J.~Materzynska, S.~Westphal, H.~Kim,
  V.~Haenel, I.~Fruend, P.~Yianilos, M.~Mueller-Freitag \emph{et~al.}, ``The
  "something something" video database for learning and evaluating visual
  common sense,'' in \emph{Proceedings of the IEEE international Conference on
  Computer Vision}, 2017, pp. 5842--5850.

\bibitem{kay2017kinetics}
W.~Kay, J.~Carreira, K.~Simonyan, B.~Zhang, C.~Hillier, S.~Vijayanarasimhan,
  F.~Viola, T.~Green, T.~Back, P.~Natsev \emph{et~al.}, ``The kinetics human
  action video dataset,'' \emph{arXiv preprint arXiv:1705.06950}, 2017.

\bibitem{arnab2021vivit}
A.~Arnab, M.~Dehghani, G.~Heigold, C.~Sun, M.~Lu{\v{c}}i{\'c}, and C.~Schmid,
  ``Vivit: A video vision transformer,'' in \emph{Proceedings of the IEEE/CVF
  International Conference on Computer Vision}, 2021, pp. 6836--6846.

\bibitem{haarnoja2018soft}
T.~Haarnoja, A.~Zhou, P.~Abbeel, and S.~Levine, ``Soft actor-critic: Off-policy
  maximum entropy deep reinforcement learning with a stochastic actor,'' in
  \emph{International Conference on Machine Learning}.\hskip 1em plus 0.5em
  minus 0.4em\relax PMLR, 2018, pp. 1861--1870.

\bibitem{grauman2022ego4d}
K.~Grauman, A.~Westbury, E.~Byrne, Z.~Chavis, A.~Furnari, R.~Girdhar,
  J.~Hamburger, H.~Jiang, M.~Liu, X.~Liu \emph{et~al.}, ``Ego4d: Around the
  world in 3,000 hours of egocentric video,'' in \emph{Proceedings of the
  IEEE/CVF Conference on Computer Vision and Pattern Recognition}, 2022, pp.
  18\,995--19\,012.

\end{thebibliography}


\begin{thebibliography}{1}

\bibitem{jax2018github}
James Bradbury, Roy Frostig, Peter Hawkins, Matthew~James Johnson, Chris Leary,
  Dougal Maclaurin, George Necula, Adam Paszke, Jake Vander{P}las, Skye
  Wanderman-{M}ilne, and Qiao Zhang.
\newblock {JAX}: composable transformations of {P}ython+{N}um{P}y programs.
\newblock {\em Software available from https://github.com/google/jax}, 2018.

\bibitem{dehghani2021scenic}
Mostafa Dehghani, Alexey Gritsenko, Anurag Arnab, Matthias Minderer, and
  Yi~Tay.
\newblock {Scenic}: A {JAX} library for computer vision research and beyond.
\newblock In {\em Proceedings of the IEEE/CVF Conference on Computer Vision and
  Pattern Recognition}, 2022.

\bibitem{ebert2021bridge}
Frederik Ebert, Yanlai Yang, Karl Schmeckpeper, Bernadette Bucher, Georgios
  Georgakis, Kostas Daniilidis, Chelsea Finn, and Sergey Levine.
\newblock Bridge data: Boosting generalization of robotic skills with
  cross-domain datasets.
\newblock In {\em RSS}, 2022.

\bibitem{haarnoja2018sacapps}
Tuomas Haarnoja, Aurick Zhou, Kristian Hartikainen, George Tucker, Sehoon Ha,
  Jie Tan, Vikash Kumar, Henry Zhu, Abhishek Gupta, Pieter Abbeel, and Sergey
  Levine.
\newblock Soft actor-critic algorithms and applications.
\newblock {\em arXiv preprint arXiv:1812.05905}, 2018.

\end{thebibliography}

\clearpage
\appendices
\section*{Overview of supplementary materials}
This appendix is organised as follows: in Section \ref{app:training_details}, we describe implementation details of the distance model and policy training. In Section \ref{app:spearman}, we discuss how HOLD models can be evaluated on human data. Ablations over several design choices are included in Section \ref{app:distance_model_ablations}. We estimate the increase in sample efficiency achieved with HOLD in Section \ref{app:long_sac_training}, and look into the coverage of the proposed robot tasks in SSv2 in Section \ref{app:ssv2_vs_robot_tasks}.

A supplementary video visualizing examples of distances predicted by HOLD-C as well as policy roll-outs is included on the project website\footnote{\href{https://sites.google.com/view/hold-rewards}{sites.google.com/view/hold-rewards}}. We also qualitatively evaluate the predictions on episodes from Bridge Data \citesupp{ebert2021bridge}, a diverse dataset of robot manipulation tasks recorded on real robots, and include examples in the video. The results suggest our distance model may well generalize to training manipulation policies on real robots.

\section{Training details}
\label{app:training_details}

Our distance models are implemented in JAX~\citesupp{jax2018github} using the Scenic library~\citesupp{dehghani2021scenic}. Hyperparameter settings are shown in Table \ref{tab:hold_r_hyperparameters} for the regression models and in Table \ref{tab:hold_c_hyperparameters} for time-contrastive training. 
For policy training, we reuse the implementation of SAC from \cite{schmeckpeper2020reinforcement} based on Softlearning \citesupp{haarnoja2018sacapps}. All RL hyperparameter settings are unchanged (included in Table \ref{tab:policy_hyperparameters} for reference).

\begin{table}[h]
    \centering
    \begin{tabular}{c|c|c}
         Parameter & ViViT & ResNet-50 \\
         \hline
         Epochs & 20 & 100 \\
         Base learning rate & 0.1 & 3e-4 \\
         Optimizer & Momentum & Adam \\
         Batch size & 64 & 32 \\
    \end{tabular}
    \vspace{0.1cm}
    \caption{Training hyperparameters for HOLD-R.}
    \label{tab:hold_r_hyperparameters}

\vspace{0.5cm}
    \centering
    \begin{tabular}{c|c|c}
         Parameter & ViT & ResNet-50 \\
         \hline
         Epochs & 5 & 100 \\
         \cline{2-3}
         Sequence length & \multicolumn{2}{c}{32} \\
         Base learning rate & \multicolumn{2}{c}{1e-4} \\
         Optimizer & \multicolumn{2}{c}{Adam} \\
         Batch size & \multicolumn{2}{c}{8} \\
         Margin ($m$) & \multicolumn{2}{c}{0.2} \\
         Pos. window & \multicolumn{2}{c}{0.2s} \\
         Neg. window & \multicolumn{2}{c}{0.4s}
    \end{tabular}
    \vspace{0.1cm}
    \caption{Training hyperparameters for HOLD-C.}
    \label{tab:hold_c_hyperparameters}

\vspace{0.5cm}
    \centering
    \begin{tabular}{c|c}
         Parameter & Value \\
         \hline
         Discount & 0.99 \\
         Initial exploration steps & 1000 \\
         Learning rate & 3e-4 \\
         Batch size & 256 \\
         Optimizer & Adam \\
         Gradient steps per environment step & 1 \\
    \end{tabular}
    \vspace{0.1cm}
    \caption{Training hyperparameters for policy training.}
    \label{tab:policy_hyperparameters}
\end{table}

\section{Distance model evaluation on human data}
\label{app:spearman}
To avoid evaluating every variation of HOLD in the target robot environment, it would be preferable to be able to rank and pre-select models based on their performance on held-out human data, and test only the most promising ones in robot policy training. However, it is difficult to evaluate generalization without access to robot data, and it is not straightforward to design a suitable test metric that captures both smoothness and correct ranking of states. In this section, we propose several relevant metrics.

For the regression models, in addition to the training objective mean squared error (MSE), we can also evaluate mean absolute error in time steps and in seconds. However, these metrics assume uniform progress at each time step toward task completion, and require high-scoring models to match the scale of ground truth time intervals. Using a hinge loss instead allows non-uniform progress and only penalizes out-of-order predictions:
\begin{equation}
    \mathcal{L}_{h} = \frac{\sum^N_{i=1}\sum^{T_i-1}_{j,k=1} \max(0, d(s^i_k,s^i_{T_i}) - d(s^i_j,s^i_{T_i}))\mathbb{I}[j < k]}{\sum^N_{i=1}(T_i-1)}.
\end{equation}
Another option is to not use time-based metrics at all. As explained in Section \ref{ssec:dist_learning},
it is ultimately more important for the distance models to preserve the ranking of states with respect to a goal frame than to reproduce $\delta$ in absolute terms. With the aim of maximally preserving pairwise rankings as defined in Section \ref{ssec:dist_learning}, we propose two further metrics, namely misclassification rate:
\begin{equation}
    \mathcal{L}_{miscl} = \frac{\sum^N_{i=1}\sum^{T_i-1}_{j,k=1} \mathbb{I}[d(s^i_k,s^i_{T_i}) > d(s^i_j,s^i_{T_i})]\mathbb{I}[j < k]}{\sum^N_{i=1}(T_i-1)},
\end{equation}
and Spearman correlation, i.e., the correlation between rankings assigned to each frame in the full sequence $s_{1:T_i-1}$, and the ground truth order.

\begin{table*}[h]
    \centering
    \setlength{\tabcolsep}{5pt}
    \begin{tabular}{|l l c|c|c|c|c|c|c|c|}
        \hline
        Model & Network & \# frames
        & Spearman & Misclassification rate & MSE & Mean error & Hinge loss \Tstrut\Bstrut\\
        \hline
         HOLD-R & ViViT & 3 & 0.6709 & 0.4539 & 499.2 & 18.0 (1.54 s) & 0.0663 \Tstrut\\
         HOLD-R & ResNet-50 & 1 & 0.7136 & \textbf{0.3976} & \textbf{482.1} & \textbf{17.3 (1.48 s)} & \textbf{0.0233} \\  
         HOLD-R & ResNet-50 & 3 & \textbf{0.7139} & 0.4385 & 514.1 & 18.1 (1.55 s) & 0.0611 \Bstrut\\ 
        \hline
         HOLD-C & ResNet-50 & 1 & 0.6246 & 0.4015 \Tstrut\\
         HOLD-C & ViT & 1 & 0.6559 & 0.4006 \Bstrut\\
        \cline{1-5}
    \end{tabular}
    \vspace{0.2cm}
    \caption{Evaluation scores on the Something-Something v2 validation set. Time-based metrics are only defined for HOLD-R as HOLD-C models do not predict time.}
    \label{tab:spearman_scores}
\end{table*}

The scores of each of the models we present are shown in Table \ref{tab:spearman_scores}. As we assume no access to robot data at distance training time, we use the SSv2 validation set as a proxy for model performance, and use Spearman correlation as an early stopping criterion. However, we observe that the scores on human data are not predictive of the downstream robot performance these models obtain, highlighting the difficulty of the domain transfer.


\begin{figure*}[h]
    \centering
    \begin{subfigure}[t]{0.4\textwidth}
    \includegraphics[width=\linewidth]{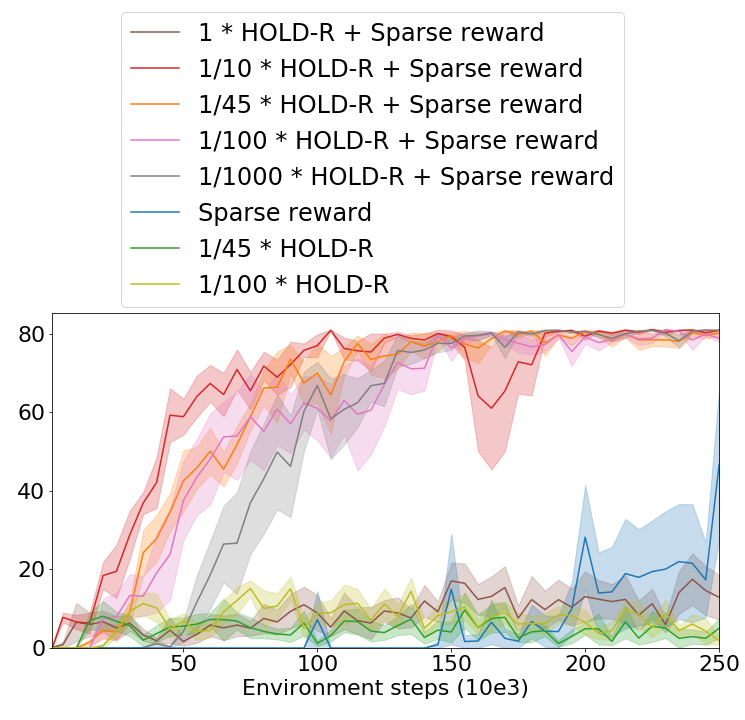}
    \caption{Effect of reward scale for HOLD-R.}
    \label{fig:holdr_scale_ablation}
    \end{subfigure}
    \begin{subfigure}[t]{0.4\textwidth}
    \includegraphics[width=\linewidth]{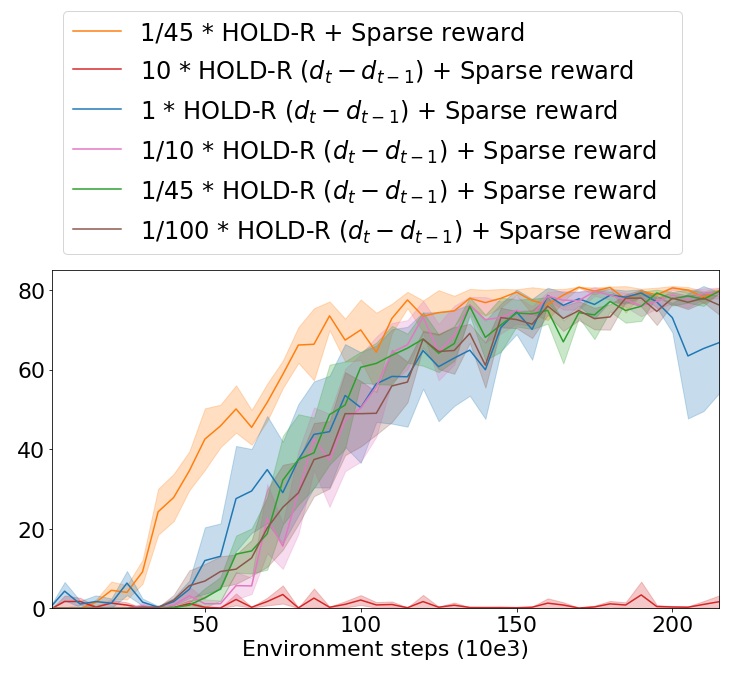}
    \caption{Cumulative distance vs. subtracting the previous distance for HOLD-R.}
    \label{fig:holdr_subprev_ablation}
    \end{subfigure}\\
    \vspace{0.1cm}
    \begin{subfigure}[t]{0.4\textwidth}
    \includegraphics[width=\linewidth]{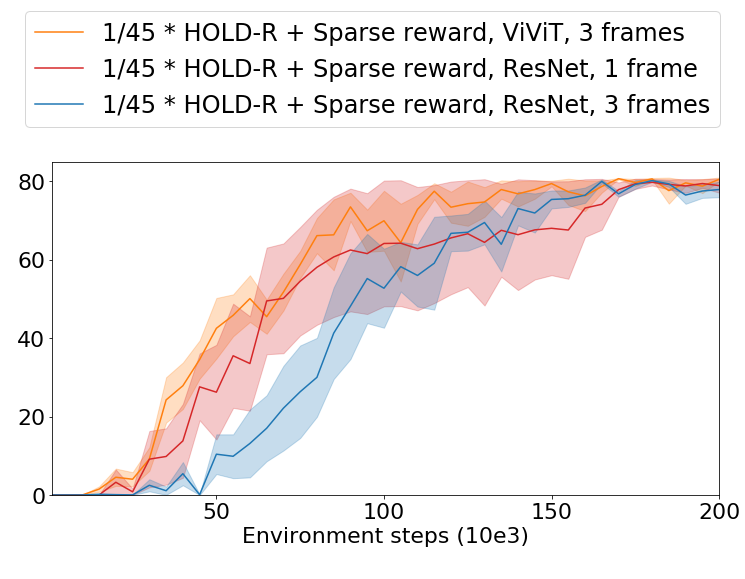}
    \caption{Effect of architecture choice for HOLD-R.}
    \label{fig:holdr_archi_ablation}
    \end{subfigure}
    \begin{subfigure}[t]{0.4\textwidth}
    \includegraphics[width=\linewidth]{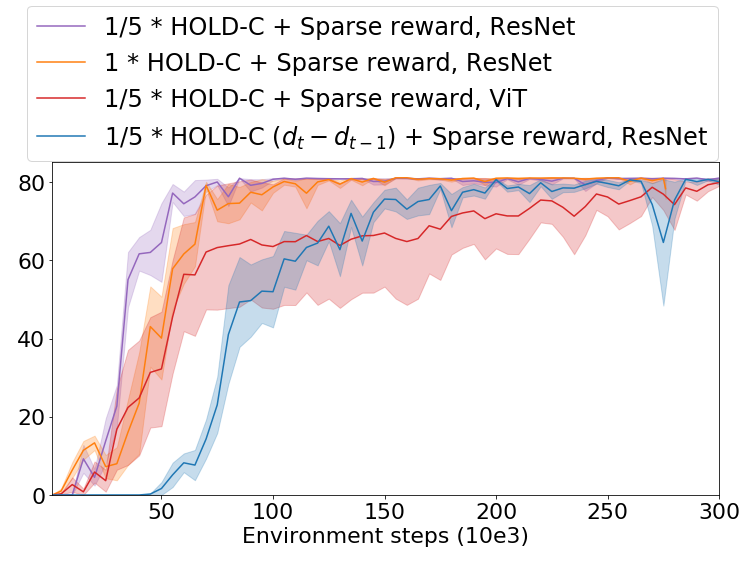}
    \caption{Effect of reward definition, scaling and architecture choice for HOLD-C.}
    \label{fig:holdc_ablation}
    \end{subfigure}
    \caption{HOLD model ablations on RLV Pushing.}
    \label{fig:distance_model_ablations}
\end{figure*}

\begin{figure}[t]
    \centering
    \begin{subfigure}[t]{0.4\textwidth}
    \includegraphics[width=\linewidth]{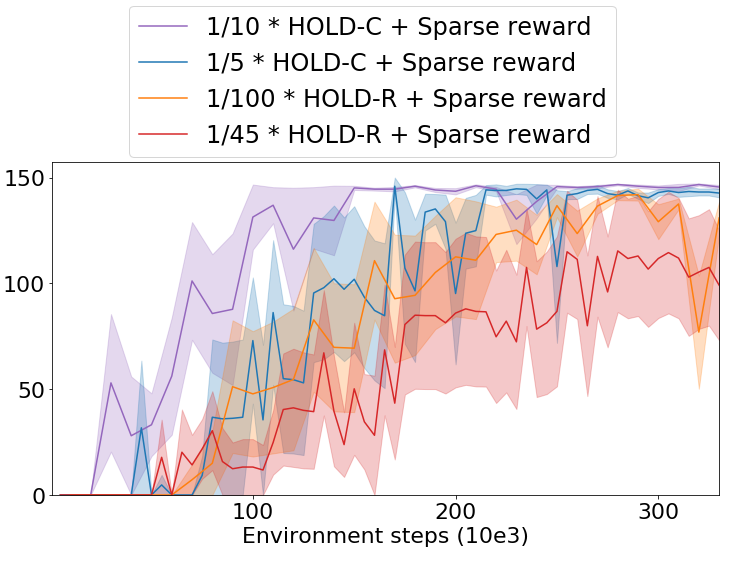}
    \end{subfigure}
    \caption{Effect of reward scale on RLV Open Drawer.}
    \label{fig:distance_model_ablations_drawer}
\end{figure}

\section{Distance model ablations}
\label{app:distance_model_ablations}
We evaluate a variety of design choices in HOLD models on the RLV Pushing task. Specifically, we compare several values for the reward normalizer $T$ introduced in Eq. (\ref{eq:func_dist_reward}), as well as variants of the network architecture and the form of the reward (cumulative vs. instantaneous, as described in Section \ref{ssec:policy_learning}).

The effect of reward scale for HOLD-R is shown in Fig. \ref{fig:holdr_scale_ablation}. While $T=10$ increases return the fastest, its performance is less stable towards the end of training and it suffers a momentary drop in performance when the policy appears to overfit to the distance reward over the sparse task reward. The normalizer $T=45$, equal to the average length of a training video in SSv2, provides the best trade-off in sample efficiency and stability and we therefore report results using this setting in Section \ref{ssec:policy_learning_results}. This value is used for all tasks except RLV Open Drawer, where the task horizon is twice as long and we found $T=100$ to work significantly better (see Fig. \ref{fig:distance_model_ablations_drawer}). To avoid further extensive tuning of the reward scale per reward model, we simply set $T$ such that the scale of initial (starting state) predictions is approximately $1/3$, the same scale as HOLD-R with $T=45$ for RLV Pushing and DVD tasks. Using this strategy, we obtain $T=5$ for HOLD-C and $T=30$ for the L2 baseline (or $T=10$ and $T=100$ for RLV Drawer, respectively), which we indeed found to perform better than respective alternatives $T=1$ and $T=45$.

As for reward definition, the cumulative distance reward clearly outperforms instantaneous distance reward (i.e. subtracting the distance at the previous time step) for both HOLD-R (Fig. \ref{fig:holdr_subprev_ablation}) and HOLD-C (Fig. \ref{fig:holdc_ablation}). Although the scale of $d_t - d_{t-1}$ is expected to be different from the scale of $d_t$ and hence the normalizer $T$ may need to be set differently, we found the HOLD-R instantaneous distance form to perform consistently worse for a wide range of values of $T$: \{0.1, 1, 10, 45, 100\}. Moreover, the choice of $T$ seemed to have very little effect on the learning performance for $T \leq 1$.

Finally, in Fig. \ref{fig:holdr_archi_ablation}, we compare the HOLD-R ViViT model against ResNet-50 conditioned on either 1 or 3 frames, but found that these smaller models had slightly worse sample efficiency in RL training than ViViT. For HOLD-C models, we found ResNet to outperform ViT, however, ViT may have benefited from longer training. In Section \ref{ssec:policy_learning_results}, we therefore report HOLD-C results using the ResNet architecture.



\begin{figure}[b!]
    \centering
    \begin{subfigure}[t]{0.23\textwidth}
    \includegraphics[width=\linewidth]{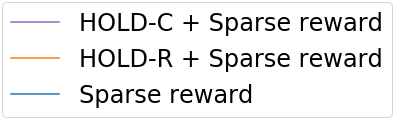}
    \end{subfigure}
    \centering
    \begin{subfigure}[b]{0.4\textwidth}
    \includegraphics[width=\linewidth]{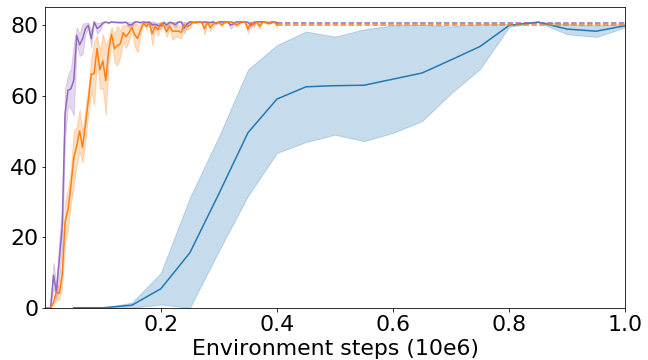}
    \caption{Pushing}
    \label{fig:pushing_sparse_long}
    \end{subfigure}
    \centering
    \begin{subfigure}[b]{0.4\textwidth}
    \includegraphics[width=\linewidth]{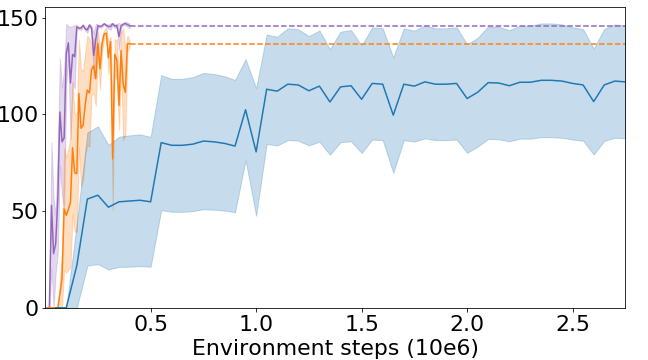}
    \caption{Drawer Opening}
    \label{fig:drawer_sparse_long}
    \end{subfigure}
    \caption{Eventual return for the sparse reward on the RLV tasks after 1--2.75 million environment steps (2.5--7x increase from Fig. \ref{fig:rlv_results}).}
    \label{fig:rlv_long_results}
\end{figure}

\section{Longer training for sparse reward}
\label{app:long_sac_training}
HOLD-C converges to a return of 80 for Pushing after 80,000 environment steps of training and a return of 145 for Drawer after 150,000 steps. In order to estimate how much training time is accelerated compared to using only the sparse reward, we also run experiments with considerably longer training for the sparse reward baseline. The return of HOLD is eventually reached after 800,000 samples for Pushing, whereas for Drawer, it is not reached even after 2.75 million steps. 
We therefore obtain a speedup of 10x for Pushing (Fig. \ref{fig:pushing_sparse_long}) and at least 18x for Drawer (Fig. \ref{fig:drawer_sparse_long}).

\section{Training data coverage of evaluated robot tasks}
\label{app:ssv2_vs_robot_tasks}

\begin{table*}[h!]
    \centering
    \begin{tabular}{|l|c|c|c|c|c|}
         \hline
         Robot task & Closest action & Closest object & \emph{a} & \emph{o} & \emph{(a, o)} \Tstrut\Bstrut\\
         \hline
         Pushing & Moving something towards the camera & cap & 927 & 611 & 3 \Tstrut\\
         Open Drawer & Opening something & drawer & 1585 & 619 & 67 \\
         Close Drawer & Closing something & drawer & 1296 & 619 & 62 \\
         Push Cup Forward & Moving something away from the camera & mug & 937 & 951 & 4\\
         Turn Faucet Right & Pushing something from left to right & faucet & 3199 & 28 & 0 \Bstrut\\
         \hline
    \end{tabular}
    \vspace{0.3cm}
    \caption{SSv2 training examples most closely matching the evaluated robot tasks. Column 2 corresponds to the most similar action template \emph{a} to each robot task, whereas column 3 lists the most similar object \emph{o} among the objects manipulated across all tasks templates. In columns 3--5, we give the number of videos in the training and validation sets labeled with either \emph{a}, \emph{o}, or both, respectively. The full dataset consists of 220,847 videos: 168,913 in the train set, 24,777 in the validation set and the remaining 27,157 in the test set.}
    \label{tab:closest_ssv2_tasks}
\end{table*}

Our distance models are not specialized for any specific task and can therefore be applied on previously unseen manipulation tasks, or tasks with very few human demonstrations. In this section, we investigate to what extent the robot tasks we evaluate on are covered in SSv2 training data. Note that our models never observe the task labels and are only trained on ungrouped SSv2 videos.

\addtolength{\textheight}{-2cm}

The tasks included in SSv2 are intentionally very diverse. As task templates include generic movements such as "Moving something up", "Moving something down", "Pushing something from left to right", "Pushing something from right to left", it is genuinely difficult to find manipulation tasks unrelated to any of the 174 templates. However, several tasks include drastically different manipulations depending on the objects considered: e.g., opening the screw cap of a bottle and opening a book use the same template "Opening something" but very different motions. As shown in Table \ref{tab:closest_ssv2_tasks}, the action-object pairs we evaluate in the robot tasks have never been demonstrated for Turn Faucet Right, and have been demonstrated $<$5 times for Push Cup Forward and RLV Pushing (moving a cap toward the camera). The objects "puck" or "disk" do not appear in SSv2, so we replace "puck" by "cap".

Moreover, on closer inspection of the three videos labeled "Moving cap towards the camera", we observe significant variation in the interpretation of the action labels themselves. Instead of pushing an object along a surface like in our robot task, one out of three videos in fact shows pulling a (baseball) cap along a surface, and the remaining two show a person holding a bottle cap and moving it directly towards the camera lens without using a surface at all. The same two motions are demonstrated for related objects such as "lid" (2 videos) and "bottle cap" (1 video). It seems unlikely that the same task, \emph{pushing} a puck towards the camera is demonstrated at all (the pushing templates featuring horizontal directions only). Similarly, only 1/4 videos labeled "Moving mug away from the camera" and 1/4 videos labeled "Moving cup away from the camera" push an object along a surface at all, and the rest perform the maneuver in the air.

We conclude that we have shown generalization to at least one and possibly multiple novel tasks which were not included in the training dataset.


\bibliographystylesupp{plain}
\bibliographysupp{root}

\end{document}